\documentclass{article}
\usepackage{spconf,amsmath,graphicx}

\usepackage{stfloats}

\usepackage{amsfonts}
\usepackage{algorithm} %format of the algorithm

\usepackage{url}
\usepackage{fixltx2e}
\usepackage{amsmath}
\usepackage{algorithmic}
\usepackage{array}
\usepackage{ifpdf}

\usepackage{amssymb}
\usepackage{multirow}
\usepackage{verbatim}

\title{Graph-Theoretic Spatiotemporal Context Modeling \\ for Video Saliency Detection}
\name{Lina Wei, Fangfang Wang, Xi Li$^{*}$\thanks{$^{*}$ Corresponding author}, Fei Wu, Jun Xiao\thanks{This work was supported in part by the National Natural Science Foundation of China under Grant U1509206 and Grant 61472353, in part by the Alibaba-Zhejiang University Joint Institute of Frontier Technologies.}}
\address{Zhejiang University\\
         College of Computer Science and Technology\\
         Hangzhou 310027, China}
\begin{document}
%\ninept
%
\maketitle
\begin{abstract}
As an important and challenging problem in computer vision, video saliency detection is typically cast as a spatiotemporal context modeling problem over consecutive frames. As a result, a key issue in video saliency detection is how to effectively capture the intrinsical properties of atomic video structures as well as their associated contextual interactions along the spatial and temporal dimensions. Motivated by this observation, we propose a graph-theoretic video saliency detection approach based on adaptive video structure discovery, which is carried out within a spatiotemporal atomic graph. Through graph-based manifold propagation, the proposed approach is capable of effectively modeling the semantically contextual interactions among atomic video structures for saliency detection while preserving spatial smoothness and temporal consistency. Experiments demonstrate the effectiveness of the proposed approach over several benchmark datasets.
\end{abstract}
\section{Introduction}
\label{sec:intro}

Video saliency detection~\cite{zhang2009sunday,MotionLS2012,3DVideo2010,Benchmark2016} is an important problem in the field of video analysis and has a wide range of applications such as object tracking, video retrieval, abnormal event detection and so on. This problem aims to automatically discover the visually interesting regions in a video sequence. In this area, a challenging issue is how to extract the atomic video structures(defined as the basic operating units which consider the spatial smoothness and temporal consistency) which reflect the intrinsical properties and the interactions among atomic units through both spatial and temporal dimensions in a unified framework. Therefore, the focus of video saliency detection is on effectively modeling the spatiotemporal relationship of the video.

In recent years, video saliency detection is typically characterized by exploring spatial and temporal properties. On one hand, the spatial saliency properties of a video sequence can be represented in multiple aspects. For example, some approaches measure saliency by local or global center-surround contrast which employs visual features such as color, intensity and orientation~\cite{itti1998model,klein2011center,GS,Lia2015,BSCA2015,Qi2017,ZhangBS2016,HanGESS2016}. With the development of convolutional neural networks, deep learning is widely used in saliency detection tasks~\cite{krizhevsky2012imagenet, li2015visual,Simonyan2015Deep,DRFI,WangRFCNN2016}. These data-driven saliency models aim to directly capture the features which are able to represent the semantic properties of salient regions by means of supervised learning from a collection of training data with saliency annotations. Furthermore, the approaches~\cite{GMR,Hypergraph} formulate saliency detection problem as a labeling task on the graph to represent the coherence and consistency among salient regions and the difference between salient regions and the background.
On the other hand, video sequence understanding also needs to take temporal properties into consideration~\cite{kim2015spatiotemporal}. Some approaches~\cite{zhai2006visual,Li2009entropy} extend existing spatial saliency detection methods by adding the temporal dimension to obtain final video saliency maps.
In essence, the task of video saliency detection is related to both spatial and temporal properties simultaneously. Therefore, how to effectively model these factors in a unified framework is the key to better explore the intrinsical properties of atomic video structures and their associated contextual interactions.

Motivated by these observations, we propose a graph-theoretic video saliency detection approach based on adaptive video structure discovery, which is carried out within a spatiotemporal atomic graph. Considering the spatial smoothness and temporal consistency, we discover atomic units to better model the intrinsical properties of a video sequence. In order to better represent the extensive information of atomic units (e.g., color, context and semantics), we combine low-level color features with highly abstract deep features. Then, we utilize the graph-based manifold propagation to model the semantically contextual interactions within atomic video structures for saliency detection while preserving spatial smoothness and temporal consistency.

The main contributions of our paper are summarized as follows:

\begin{figure*}[h]
\begin{center}
\includegraphics[width=17cm]{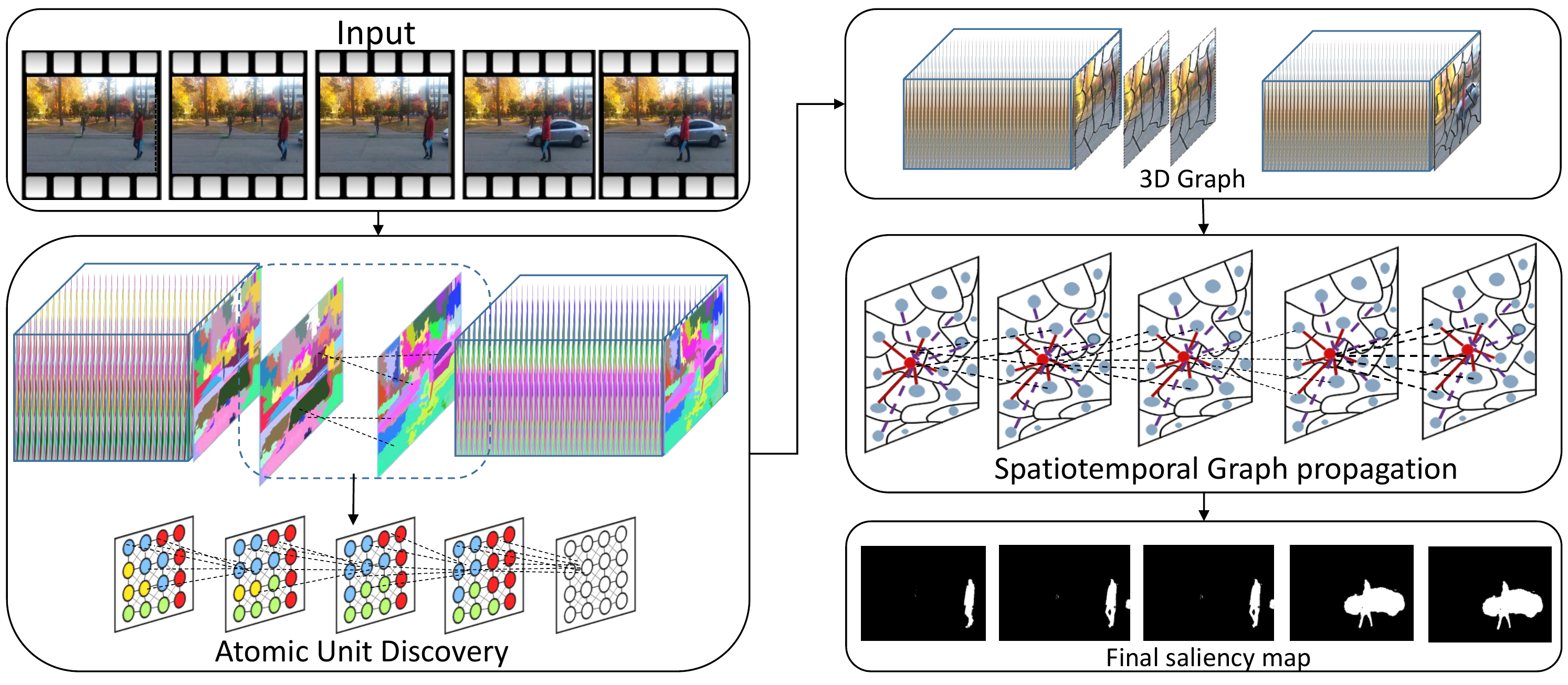}
\end{center}
\vspace{-1.5em}
\caption[]{Illustration of the proposed approach for video saliency detection. First, an adaptive graph-based atomic structure is discovered according to the spatiotemporal segmentation. Second, construct a spatiotemporal graph. Third, utilize the graph-based manifold propagation to model the semantically contextual interactions within atomic video structures.}
\vspace{-1.5em}
\label{fig2}
\end{figure*}

1) We propose a graph-based video saliency detection approach based on adaptive video structure discovery, which is carried out in a spatiotemporal atomic graph. As the representation units of a video, the atomics are able to model both spatial and temporal layouts of the video in a unified structure.

2) In the spatiotemporal context model, we utilize graph propagation for video saliency detection. Through graph-based manifold propagation, the proposed approach is capable of effectively modeling the semantically contextual interactions within atomic video structures for saliency detection while preserving spatial smoothness and temporal consistency. Therefore, our work effectively explores the intrinsic problem on how to make the connection from spatio-temporal  context modeling to video saliency, resulting in promising saliency detection results.

\section{Our Approach}
\label{sec:Approach}

\subsection{Problem Formulation}
\label{ssec:PF}
In this work, we aim to build a framework to automatically solve the video saliency detection problem. We think of a t-frame video as a space-time lattice $\mathcal{F} \doteq \Lambda^{2} \times \mathbb{Z}$ where $\Lambda^{2}$ denotes the 2D frame of the video, and $\mathbb{Z}$ the time axis, so $\mathcal{F}$ can be viewed as a 3D matrix of size $m\times n\times t$. Our goal is to discover the salient region $\mathcal{R} \in \mathcal{F}$ of a video so that $\mathcal{F}[ i,j,k ]\in \mathcal{R}$ if and only if the pixel in location  $(i,j,k)$ is salient. Finally, the framework outputs salient maps $M$ for video sequence which defined as a set of frame saliency maps. The video salient maps not only segment the visually interesting regions of each frame, but also capture the motion regions which contain temporal consistency and the abrupt changes during the continuous frames.

A video is substantially a temporally adjacent sequence which encodes both static and dynamic information of a scene. The pixels within a spatiotemporal local region possess the property of spatial smoothness. Meanwhile, the motion trends of its contents preserve temporal smoothness across consecutive frames.

Based on these observations, we discover the spatiotemporal atomic units of a video to exploit consistency in time and smoothness in space. The video becomes a structural layout of atomic units which contain rich information about its appearance, motion trends, semantic interactions and so on. We cultivate these information by representing the units with combinatorial pixel-wise features.

To define the contextual interactions of the video, a compact 3D adjacency graph $G=(V,E)$ is constructed with nodes $V$ (the atomic units) and edges $E$ (the interaction among units) along the spatial and temporal dimensions. Then, we exploit the saliency distribution of the video by means of graph-based manifold propagation. The salient region $\mathcal{R}$ of this video is thus defined as the subgraph activated by high saliency scores during the propagation procedure.
Figure~\ref{fig2} shows the pipeline of our method.

\subsection{Graph-based Modeling}
\label{ssec:GbM}

\subsubsection{Spatiotemporal Atomic Unit Discovery}
\label{sssec:SAUD}
The core idea of our framework is to exploit the spatial smoothness and temporal consistency of a video simultaneously. These properties are enclosed in the interactions among pixels within a spatiotemporal atomic unit. So the discovery of atomic units is converted to a pixel clustering problem in accordance with coherence in appearance and motion trend. This problem can be solved by spatiotemporal segmentation~\cite{XuWhCoICCV2013}.

For a given video $\mathcal{F}$, our goal is to discover its atomic structure $\mathcal{S}$. In order to obtain the atomic units of the most suitable granularity, the segmentation is conducted in a hierarchical way resulting in $m$ layers of individual segments $\mathcal{S} \doteq\{S^{1}, S^{2}, ..., S^{m}\}$ where each layer $S^{i}$ is a set of atomic units
$\{s_{1}, s_{2},...\}$
such that $\bigcup_{j}s_{j} = \mathcal{F}$ and
$s_{i}\bigcap s_{j} = \emptyset$
for all pairs of atomic units.

The spatiotemporal atomic units contain extensive information about the video such as color, context, semantics and so on. So we represent them with LAB color features~\cite{FSR} and FCN features~\cite{FCN} in a pixel-wise manner. As the atomic units are of arbitrary shapes and sizes, we aggregate the pixel-wise features by regional pooling.

Therefore, two feature maps $F_{LAB}, F_{FCN} \in \mathbb{R}^{m\times n\times t}$ are extracted from the video. For the i-th atomic unit $s_{i}$, its feature vectors aggregated by regional pooling are denoted as $f_{i}^{LAB}$ and $f_{i}^{FCN}$. The video can be transformed to $\mathcal{F} \doteq {[ f_{1},...,f_{N} ]}^{T}$, where N is the total number of atomic units generated after discovery.

\subsubsection{Graph-based Context Model}
\label{sssec:GbCM}
The spatial smoothness and temporal consistency of a video also lie in the contextual interactions among the atomic units.

To address this problem, we construct a spatiotemporal adjacency graph $G = (V, E)$ to model the contextual interaction among atomic units, where $V$ is a set of atomic units denoted as $\{f_{i}\}^{N}_{i=1}$, and $E$ is a set of undirected edges which connect neighboring atomic units. Here, the spatiotemporal adjacency is defined as the relationship that atomic units are neighboring in both spatial and temporal dimensions. Therefore, the atomic unit level graph with affinity matrix $W = (w_{ij})_{N\times N}$ is constructed as follows:
\vspace{-0.2cm}
\begin{equation}\label{eq:W}
w_{ij}=
\left\{
\begin{array}{ll}
K(f_{i},f_{j}) & \mbox{if} \thinspace f_{i} \thinspace \mbox{and} \thinspace {f_{j}} \thinspace \mbox{with spatiotemporal adjacency}
\vspace{0.3cm}\\
0 & \mbox{otherwise}
\end{array}
\right.
\end{equation}
where $K(f_{i}, f_{j})$ stands for the RBF kernel evaluating the feature similarity (such that $K(f_{i},f_{j}) = \exp(-\frac{1}{\rho}\|f_{i} - f_{j}\|_{2}^{2})$ with $\rho$ being a scaling factor). The weights are computed on the basis of the distance in the feature space, and in our case, two different 3D graphs are constructed in LAB feature space and FCN feature space respectively.

\begin{algorithm}[t]
\renewcommand{\algorithmicrequire}{\textbf{Input:}}
\renewcommand{\algorithmicensure}{\textbf{Output:}}
\renewcommand{\algorithmicreturn}{\textbf{Return:}}
\caption{Video saliency detection}
\label{alg1}
\begin{algorithmic}
\REQUIRE Video sequence $\mathcal{F}$
\ENSURE Video saliency map $M$

\STATE
%\tcc{Spatiotemporal atomic unit discovery}
    1. Given video $\mathcal{F}$, discover the atomic structure $\mathcal{S}$;
\STATE
    2. Represent the atomic units with aggregated pixel-wise features by regional pooling ${[ f_{1},...,f_{N} ]}^{T}$ over the LAB and FCN feature spaces;
%\tcc{The structured learning part}
\STATE
    3. Construct the spatiotemporal adjacency graph $G = (V, E)$ with an affinity matrix according to Eq.~\eqref{eq:W};
\STATE
    4. Compute final saliency scores $g\ast$ according to Eq.~\eqref{eq:g1} and generate two saliency maps $M_{LAB}$ and $M_{FCN}$ respectively;
\STATE
    5. Combine $M_{LAB}$ and $M_{FCN}$ to generate the final map $M$ according to Eq.~\eqref{eq:SFCN}
\RETURN Video saliency map $M$

\end{algorithmic}
\end{algorithm}

\subsection{Saliency Propagation}
\label{ssec:SP}
We exploit the long range contextual correlation of the graph model which defines the saliency distribution of the video by graph-based propagation. The salient region of the video is thus a subgraph of its 3D graph model activated from saliency propagation.

In this framework, propagation is conducted in the manner of manifold ranking~\cite{GMR}. Let
$g: X\rightarrow \mathbb{R}^{n}$
denote a ranking function, $g=[g_{1}, g_{2},..., g_{n}]^{T}$, $g_{i}$ is the saliency score of atomic unit $f_{i}$. Let $q=[q_{1}, q_{2},..., q_{n}]^{T}$ be an initial saliency vector obtained through coarse-grained foreground segmentation from standard computer vision techniques (e.g., background subtraction or object detection). The graph $G$ is associated with an edge-weighted matrix $W$ with its degree matrix being $D = diag{\{d_{11},..., d_{nn}\}}$, where $d_{ii} = \sum _{j}w_{ij}$. Hence, our task is to learn an energy function within the following optimization framework:

\vspace{-0.3cm}
\begin{equation}\label{eq:g}
g\ast = \arg \min_{g} \frac{1}{2}(\sum^{n}_{i,j=1}w_{ij}\|\frac{g_{i}}{\sqrt{d_{ii}}}-\frac{g_{j}}{\sqrt{d_{jj}}}\|_{2}^{2}+\mu \sum^{n}_{i=1}\|g_{i}-q_{i}\|_{2}^{2})
\end{equation}
%\vspace{-0.3cm}
where $\mu$ is the smoothing factor. The closed form solution of Eq.~\eqref{eq:g} is:
\vspace{-0.3cm}
\begin{equation}\label{eq:g1}
g\ast = (I- \frac{1}{1+\mu} L)^{-1}q
\end{equation}
where $I$ is an identity matrix and $L$ is the normalized Laplacian matrix for the affinity matrix $W$, $L= D^{-1/2}WD^{-1/2}$.

%Similar to~\cite{Zhou2004Ranking}, we utilize the approximated unnormalized Laplacian matrix to obtain the saliency score for better performance.

%\begin{equation}\label{eq:g2}
%g\ast \doteq (D- \frac{1}{1+\mu} W)^{-1}q
%\end{equation}

For the LAB and FCN feature spaces, we propagate the saliency map denoted as $M_{LAB}$ and $M_{FCN}$ respectively. The final saliency map of the video is then computed as:
\vspace{-0.2cm}
\begin{equation}\label{eq:SFCN}
\begin{split}
 M = M_{LAB}^{1-\beta}  \circ   M_{FCN}^{\beta}
\end{split}
\end{equation}
where $\beta$ is a trade-off control factor and $\circ$ is the element-wise multiplication operator. The whole video saliency detection algorithm is summarized in Algorithm 1.

\section{Experiments}
\label{sec:Exp}
\subsection{Datasets and Evaluation Measures}
\label{ssec:DaEM}
In this section, we evaluate the performance of the proposed algorithm on 10 test video sequences taken by static camera from MPEG dataset, NTT dataset and MCL dataset~\cite{Akamine2012,kim2015spatiotemporal}. These three datasets\footnote{The details of these datdsets are shown in the supplementary material} include outdoor and indoor scenes containing several challenging conditions such as low resolution, light changes, abrupt changes, occlusion, multi-objects, high scene complexity and so on.
%(shown in Table~\ref{tab:dataset}).

We also utilize the AUC score which calculates the area under the receiver operating characteristic curve (ROC) to show the relationship between the true positive rate (TPR) and the false positive rate (FPR). Another evaluation measure is the normalized scan-path saliency (NSS) score~\cite{CBGA2005}.

%\begin{table}[t] \scriptsize %\tiny%\footnotesize
% \caption{Properties of test video sequences}
%\renewcommand{\tabcolsep}{2pt}
%\centering

%\resizebox{1\linewidth}{!}{
 %   \begin{tabular}{|c|c|c|c|c|c|c|c|c|c|} \hline
%\multirow{2}[0]{*}{Dataset} & Video & Scene & Objects & Video  & Scene  & Light \\
%                  & Sequence & Type & Num & Resolution & Complexity  & Changes \\
%    \hline
%\multirow{1}[0]{*}{MPEG} & Hall1  & Indoor & Multi & High  & Mid  & Light\\
%    \hline
%\multirow{3}[0]{*}{NTT} & Bird1  & Outdoor & Single  & High  & High  & Severe \\
%          & Bird2  & Outdoor & Single & Low  & High  & Severe \\
%          & Horse   & Outdoor & Multi & Low  & Mid  & Severe \\
%    \hline
%\multirow{6}[0]{*}{MCL} & Car  & Outdoor & Multi & Low  & High  & Severe \\
%          & Campus  & Outdoor & Multi & Low  & High  & Severe \\
%          & Crowd   & Outdoor & Multi & High  & High  & Severe \\
%          & Road    & Outdoor & Multi & Low  & High  & Severe \\
%         & Square  & Indoor  & Multi & High  & High  & Light \\
%        & Stair   & Outdoor & Multi & High  & High  & Severe \\
%    \hline
%     \end{tabular}%
%}
%   \vspace{-1.5\baselineskip}
%\vspace{-1.5em}
%\label{tab:dataset}%
%\end{table}%add * to use two columns

\subsection{Implementation Details}
\label{ssec:ID}
During the atomic units discovery procedure, the segmentation is carried out by the GBH method~\cite{XuCoCVPR2012,XuWhCoICCV2013} with the streaming level set to 10. We represent every pixel with a $3$ dimensional LAB feature vector and a $512$ dimensional FCN feature vector. The FCN features are extracted with FCN-32s network, implemented on the basis of the Caffe~\cite{Caffe2014,li2015visual} toolbox. For the feature representation, we conduct intra-frame max-pooling and inter-frame average-pooling within atomic units. In saliency propagation, there are three parameters in the proposed algorithm, the scalling factor $\rho$ which controls the strength of the weight in the RBF kernel is set to $0.1$.  The smoothing factor $\mu$ in Eq.~\eqref{eq:g} is set to $0.1$. The fusion weight $\beta$ in Eq.~\eqref{eq:SFCN} is set to $0.7$. All the above parameters are fixed throughout all the experiments.
%\begin{figure}[h]
%\begin{center}
%\includegraphics[width=1\linewidth]{res/InitialCom.pdf}
%\end{center}
%\vspace{-1.5em}
%\caption[]{Illustration of graph propagation. (a) shows the input frame. (b) is the visualization of the initial saliency values and (c) is the resulting saliency map generated by the proposed approach. Clearly, our approach obtains more visually feasible results.}
%\vspace{-1.5em}
%\label{fig:InitialCom}
%\end{figure}

%\vspace{4.5em}

\begin{figure}[t]
\begin{center}
\includegraphics[width=1\linewidth]{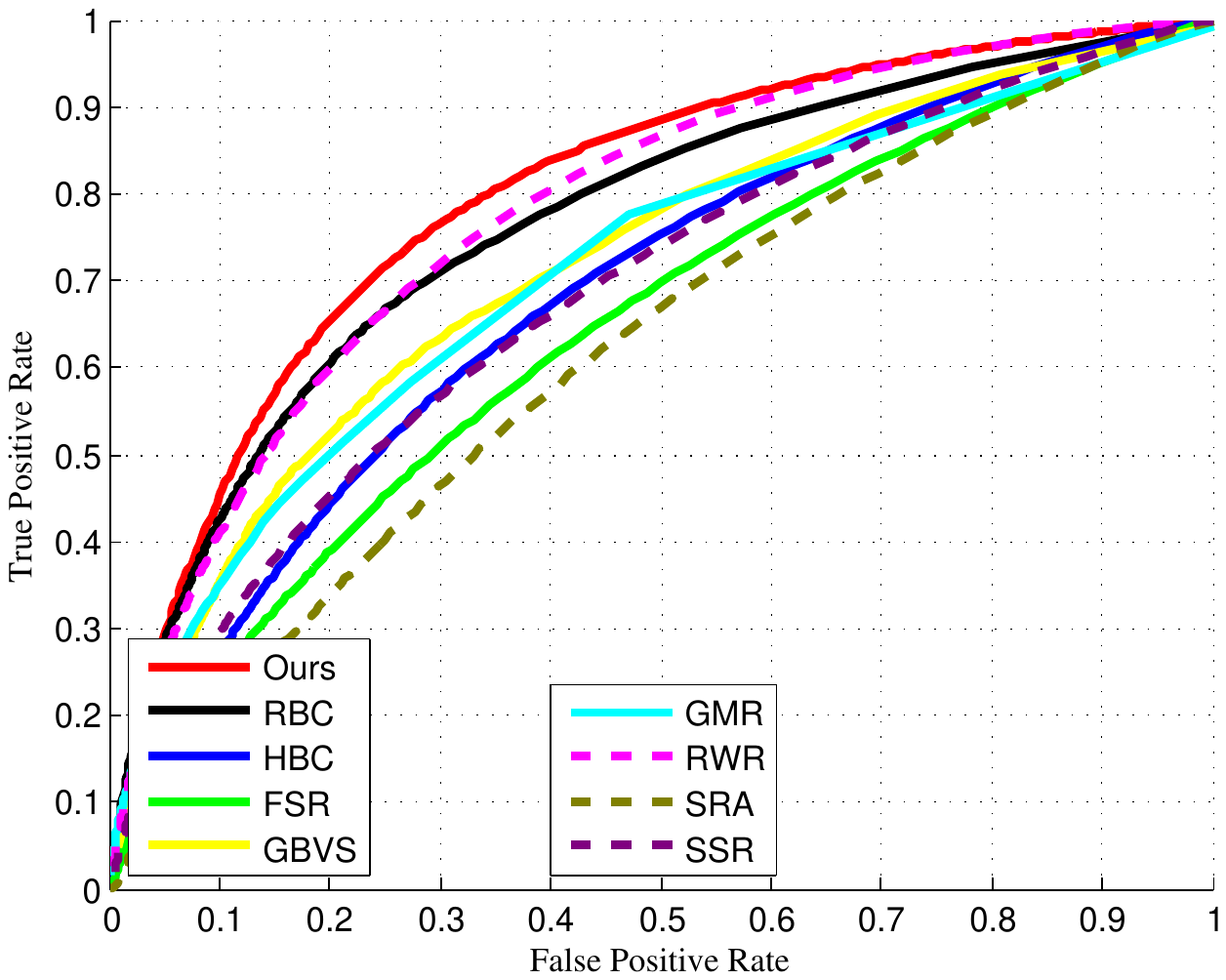}
\end{center}
\vspace{-1.5em}
\caption[]{The quantitative ROC results of all the comparison approaches under the average over all datasets. Clearly, our approach achieves the best performance over the competing approaches.}
\vspace{-1.5em}
\label{fig:ROC}
\end{figure}

\subsection{Analysis of Proposed Approach}
\label{ssec:AoPA}

\begin{figure}[h]
\begin{center}
\includegraphics[width=1\linewidth]{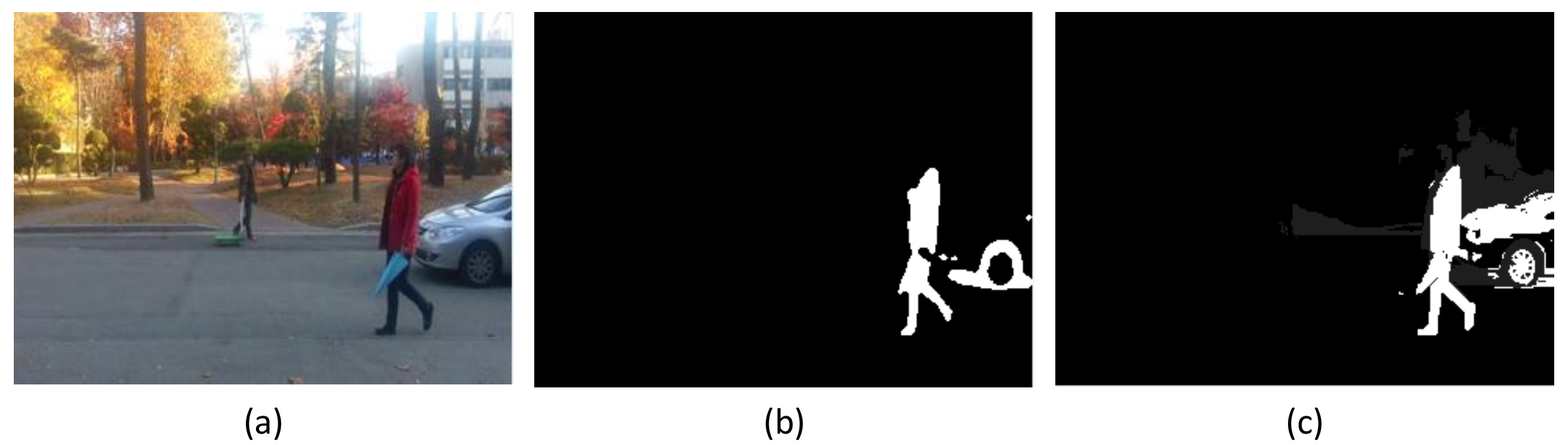}
\end{center}
%\vspace{-1.5em}
\caption[]{Illustration of graph propagation. (a) input frame. (b) the visualization of the initial saliency values and (c) the resulting saliency map.}
%\vspace{-1.5em}
\label{fig:InitialCom}
\end{figure}

We quantitatively and qualitatively compare the proposed approach with several comparison methods including GBVS~\cite{GBVS}, SRA~\cite{SRA}, SSR~\cite{MotionLS2012}, RWR~\cite{kim2015spatiotemporal}, FSR~\cite{FSR}, HBC~\cite{ChengPAMI15}, RBC~\cite{ChengPAMI15} and GMR~\cite{GMR}. The experimental results\footnote{More qualitative and quantitative results can be found in the supplementary material} show that our approach is able to effectively capture the whole object structure information by graph-based context modeling. Figure~\ref{fig:ROC} shows the corresponding ROC curves of all the comparing approaches on the average result of all testing video sequences. From Figure~\ref{fig:ROC}, we observe that the proposed approach achieves a better performance than the other ones. Table~\ref{tab:AUC} report the quantitative saliency detection performance on the AUC. It is clearly seen that our approach performs better against the comparison methods. Figure~\ref{fig:InitialCom} shows an example of the experimental results for graph propagation. From Figure~\ref{fig:InitialCom} (b) and (c), we observe that our approach is able to effectively capture the whole object structure information by graph-based context modeling.
%Figure~\ref{fig:fig7} shows that our approach is able to obtain more visually consistent and feasible saliency detection results than the other methods.
\begin{table}[h] \scriptsize %\tiny%\footnotesize
 \caption{Comparison of AUC scores of the proposed algorithm and other comparing methods. Our approach achieves the best performance in this metric.}
\renewcommand{\tabcolsep}{2pt}
\centering

\resizebox{1\linewidth}{!}{
    \begin{tabular}{|c|c|c|c|c|c|c|c|c|c|}
        \hline
        Dataset       & Ours & FSR & GBVS & GMR  & RWR  & HBC   & RBC   & SRA    & SSR   \\
        \hline
        \multirow{1}[0]{*}{Hall1}   & \textbf{0.8368} & 0.8136 & 0.8175 & 0.8143 & 0.8295 & 0.8154 & 0.8194 & 0.7803 & 0.8105  \\

        \hline
        \multirow{1}[0]{*}{bird1}   & \textbf{0.7981} & 0.7504 & 0.7446 & 0.7498 & 0.7686 & 0.7413 & 0.7529 & 0.7291 & 0.7501  \\

        \hline
        \multirow{1}[0]{*}{Bird2}   & \textbf{0.7741} & 0.7459 & 0.7302 & 0.7389 & 0.7691 & 0.7409 & 0.7518 & 0.7320 & 0.7530  \\

        \hline
        \multirow{1}[0]{*}{Horse}   & \textbf{0.7243} & 0.6801 & 0.6821 & 0.6917 & 0.7069 & 0.6832 & 0.6957 & 0.6810 & 0.6921  \\

        \hline
        \multirow{1}[0]{*}{Car}   & \textbf{0.7191} & 0.7023 & 0.7003 & 0.7008 & 0.7114 & 0.7001 & 0.7117 & 0.6923 & 0.7023  \\

        \hline
        \multirow{1}[0]{*}{Campus}   & \textbf{0.7238} & 0.7059 & 0.7055 & 0.7059 & 0.7183 & 0.7007 & 0.7058 & 0.6961 & 0.7061  \\

        \hline
        \multirow{1}[0]{*}{Crowd}   & \textbf{0.7713} & 0.7485 & 0.7504 & 0.7488 & 0.7673 & 0.7401 & 0.7499 & 0.7380 & 0.7498 \\

        \hline
        \multirow{1}[0]{*}{Road}   & \textbf{0.7234} & 0.7065  & 0.7062 & 0.7069 & 0.7176 & 0.7039 & 0.7071 & 0.7009 & 0.7097 \\

        \hline
        \multirow{1}[0]{*}{Square}   & \textbf{0.7217} & 0.7143 & 0.7148 & 0.7165 & 0.7157 & 0.7139 & 0.7154 & 0.7041 & 0.7035 \\

        \hline
        \multirow{1}[0]{*}{Stair} & \textbf{0.7208} & 0.7104 & 0.7128 & 0.7106 & 0.7154 & 0.7102 & 0.7121 & 0.7017 & 0.7033  \\

        \hline
    \end{tabular}%
}
%   \vspace{-1.5\baselineskip}
\label{tab:AUC}%
\end{table}%add * to use two columns

%\begin{figure}
%\begin{center}
%\includegraphics[width=1\linewidth]{res/figure7.pdf}
%\end{center}
%\vspace{-1.5em}
%\caption[]{Qualitative comparison results of different approaches over several sample frames (shown in the bottom line). Clearly, our approach obtains more visually feasible saliency detection results than the comparison approaches.}
%\vspace{-1.5em}
%\label{fig:fig7}
%\end{figure}

\vspace{-2em}
\section{Conclusion}
\label{sec:Con}

In this paper, we propose a graph-theoretic video saliency detection approach based on adaptive video structure discovery, which is carried out within a spatiotemporal atomic graph. We adopt the atomic units to better capture the intrinsical properties of video. Furthermore, the proposed approach utilize graph-based manifold propagation to model the semantically contextual interactions within atomic video structures to generate video saliency maps. We evaluate the proposed approach on $10$ video sequences and achieve promising results in comparison with $8$ state-of-the-art methods. The experimental results demonstrate that the proposed approach performs well in different evaluation metrics on complex video sequences.

%\section{REFERENCES}
%\label{sec:ref}

% References should be produced using the bibtex program from suitable
% BiBTeX files (here: strings, refs, manuals). The IEEEbib.bst bibliography
% style file from IEEE produces unsorted bibliography list.

% -------------------------------------------------------------------------
\small
\bibliographystyle{IEEEbib}
\bibliography{videosaliency}

%\bibliographystyle{videosaliency}
%\bibliography{strings,refs}

\end{document}